%% file: main.tex
\documentclass[twoside,11pt]{article}

\usepackage{mlhc} 

\usepackage{epsfig}
\usepackage{graphicx}
\usepackage{amsmath}
\usepackage{amssymb}
\usepackage{mathptmx}
\usepackage{graphicx}
\usepackage{subfigure}
\usepackage[textwidth=2cm,colorinlistoftodos]{todonotes}
\usepackage{xspace}
\usepackage{latexsym}
\usepackage{amsmath,amssymb,graphicx,xspace}
\usepackage{times}   
\usepackage{multirow}
\usepackage{footnote}

\usepackage{color,comment,rotating,subfigure,url,xspace} 
\usepackage{tikz}
\usetikzlibrary{arrows,shadows,shapes,backgrounds,decorations,snakes,fit}
\usepackage{todonotes}
\usepackage{enumerate}

\newcommand\blfootnote[1]{%
  \begingroup
  \renewcommand\thefootnote{}\footnote{#1}%
  \addtocounter{footnote}{-1}%
  \endgroup
}

\ShortHeadings{Diagnostic Prediction Using Discomfort Drawings with IBTM}{Zhang,  Kjellstr\"om, PhD, Ek, PhD, Bertilson, MD,PhD }

\begin{document}

\title{Diagnostic Prediction Using Discomfort Drawings with IBTM}

\author{\name Cheng Zhang, Hedvig Kjellstr\"om \email \{chengz, hedvig\}@kth.se\\
       \addr {Robotics, Perception and Learning (RPL), KTH Royal Institute of Technology}\\
       RPL/CAS, CSC,  KTH, 100 44 Stockholm, Sweden 
       \AND  
       \name Carl Henrik Ek \email carlhenrik.ek@bristol.ac.uk \\
        \addr {University of Bristol, UK}
       \AND
       \name Bo C. Bertilson \email bo.bertilson@ki.se \\
       \addr Department of Neurobiology, Care Sciences and Society, Karolinska Institutet\\
       Alfred Nobels allé 12, 14183 Huddinge, Stockholm, Sweden 
}

\maketitle
\blfootnote{{This research has been supported by the Swedish Research Council (VR), Stiftelsen Promobilia and Vinnova.}}

\begin{abstract}
In this paper, we explore the possibility to apply machine learning to make diagnostic predictions using discomfort drawings. A discomfort drawing is an intuitive way for patients to express discomfort and pain related symptoms. These drawings have proven to be an effective method to collect patient data and make diagnostic decisions in real-life practice. 
A dataset from real-world patient cases is collected for which medical experts provide diagnostic labels. 
Next, we use a factorized multimodal topic model, Inter-Battery Topic Model (IBTM),  to train a system that can make diagnostic predictions given an unseen discomfort drawing. The number of  output diagnostic labels is determined by using mean-shift clustering on the discomfort drawing. Experimental results show reasonable predictions of diagnostic labels given an unseen discomfort drawing. 
Additionally, we generate synthetic discomfort drawings with IBTM given a diagnostic label, which results in typical cases of symptoms.
The positive result indicates a significant potential of machine learning to be used for parts of the pain diagnostic process and to be a decision support system for physicians and other health care personnel.
\end{abstract}

\input{intro}
\input{probstate}

\input{model}

\input{exp}
\vspace{-15pt}
\input{discussion}
{
\small
\bibliography{ref}
}

\end{document}

%% file: intro.tex
\section{Introduction}
\label{sec:intro}

A discomfort drawing is a drawing on the image of a body where a patient may shade all areas of discomfort in preparation for a medical appointment. The drawing has been shown to be able to make diagnostic predictions - especially to discern neuropathic from nociceptive and psychiatric diseases  [{\color{blue}\cite{bertilson2007pain}}]. The use of drawings (pain drawing) to collect data from patients was first reported by Palmer in 1949 [{\color{blue}\cite{palmer1949pain}}] and has been studied in clinical settings showing high diagnostic predictive value especially in spine related pain by  [{\color{blue}\cite{ohnmeiss1999relation, vucetic1995pain, albeck1996critical, tanaka2006cervical}}]. The pain drawing, where different signs mark different kind of pain, is still in use at many clinics. As a more recent method, the discomfort drawing (a revised pain drawing) instructs the patient to shade all areas of discomfort. This method  may have some possible benefits compared to  pain drawings due to the fact that many different symptoms may arise from disfunction of  the same body organ and /or nerve~ [{\color{blue}\cite{bertilson2003reliability,bertilson2007pain,bertilson2010assessment}}].  
Hence, we focus on the use of  discomfort drawings. 

To find high-quality diagnostic prediction methods is a goal of health care  as well as the machine learning community. For example, machine learning for electrocardiogram (EKG) diagnostic prediction  [{\color{blue} \cite{kukar1999analysing}}] has been used for years as a decision support system for health care personal. However, the most common and most costly medical problem is unspecific pain and discomfort  [{\color{blue}\cite{upshur2010they} }] to which machine learning has not been applied yet. In this paper, we focus on applying machine learning for diagnosing pain-related problems using discomfort drawings. 

Topic models~ [{\color{blue}\cite{blei03latent}}], a type of generative models, have been successfully applied in different domains, such as information retrieval and computer vision  [{\color{blue}\cite{wang2009simultaneous, newman2006statistical, hospedales11, zhang13contextual}}]. 
With efficient inference algorithms [{\color{blue}\cite{hoffman2010online, ranganath2013adaptive}}],  these models can handle both  small and big datasets, in complete data and in incomplete scenarios.    Additionally, they are highly interpretable and can be used to generate missing data.
In our application of using discomfort drawings for diagnostic prediction, the data consist of multiple modalities (drawings and labels). 
Hence, a multi-modal topic model [{\color{blue}\cite{blei2003modeling, wang2009simultaneous,zhang16IBTM}}] is needed. Traditional multi-modal topic model [{\color{blue}\cite{blei2003modeling, zhang13contextual}}]  represent all the information contained in the data, hence these models are not robust to noise. A recent advancement in multi-modal topic models shows that Inter-Battery Topic Model (IBTM)~[{\color{blue}\cite{zhang16IBTM}}] is robust to noise in the data by explaining away irrelevant parts of the information.
Therefore, in this paper IBTM is adapted to predict diagnostic labels given a discomfort drawing. IBTM was originally proposed for representation learning and applyed for classification tasks. In this paper, we adapt the framework for diagnostic label prediction and use mean-shift clustering [{\color{blue}\cite{comaniciu2002mean}}] to determine the number of diagnostic predictions that the system needs to make.

The main contribution of this paper lies in the modification and use of IBTM for diagnostic prediction with discomfort drawings. 
This is a novel application of a principled framework. For this purpose, a dataset was collected from real-world clinical cases with medical expert labels. The experiments show that the adapted IBTM makes reasonable diagnostic predictions.  Additionally, the model also contributes to the interpretability of the data for humans and may further provide insight into the diagnostic procedure. Our approach shows that the use of machine learning in the assessment of discomfort drawings is a promising direction.

%% file: probstate.tex
\vspace{-5pt}
\section{Problem Statement}

\begin{table}[h]
\centering
\begin{tabular}{  c  p{9cm} }
\multirow{3}{*}{
\includegraphics[height=4cm]{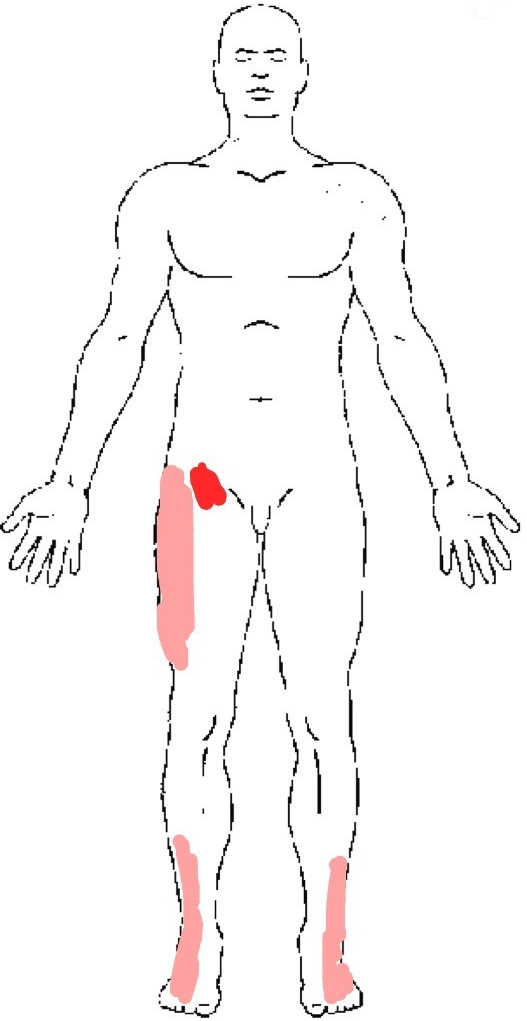}
\includegraphics[height=4cm]{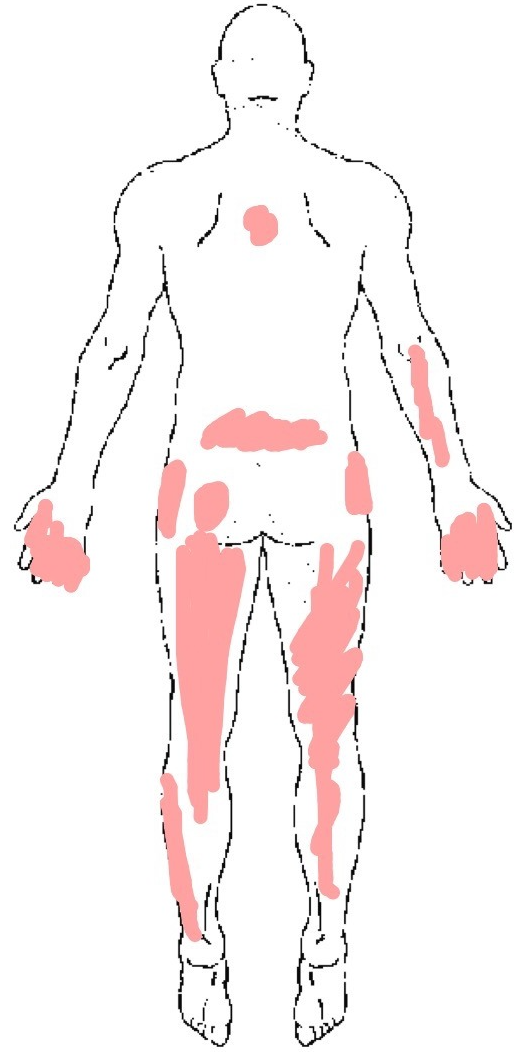}
}
&\textbf{Symptom diagnoses:}
Interscapular discomfort; R arm discomfort; B hands discomfort; Lumbago; B crest of the ilium discomfort; L side thigh discomfort; B back thigh discomfort; B calf discomfort; B achilles tendinitis; B shin discomfort; R inguinal discomfort;\\[5pt]
& \textbf{Pattern diagnoses}
B L5 Radiculopathy; B S1 Radiculopathy; B C7 Radiculopathy;\\[5pt]
& \textbf{Pathophysiological diagnoses}
DLI L4-L5; DLI S1-S2; DLI C6-C7\\
\end{tabular}
\caption{\footnotesize Discomfort drawings (left) and diagnoses by medical expert (right). R stands for right-side, L stands for left-side and B stands for bilateral.  DLI refers to discoligament injury.}
\vspace{-1cm}
\label{fig:dataExp}
\end{table}

At some clinics which treat pain-related problems, a patient is asked to shade all areas of discomfort on a drawing of a body. The intensity of shade should  indicate the level of discomfort. The patient is typically also asked to specify what type of discomfort they experience and furthermore to describe the discomfort-level over time. During a patient interview additional  information regarding symptoms, prior treatment and  experiences may be added to provide the health care personnel with sufficient information to make a diagnostic prediction that can guide the treatment.

In this paper we focus on diagnostic prediction solely based  on areas of discomfort which is the key information.
Table \ref{fig:dataExp} shows an example of discomfort drawings and their diagnoses.  
On a standard body contour the discomfort regions are marked in red. 
The right column shows the diagnostic label provided by medical experts which are roughly ordered by symptom diagnoses, possible pattern diagnoses and possible pathophysiological diagnoses. The dataset was collected in a Swedish clinic based on real-world patient cases, hence the diagnostic labels are originally given in Swedish. These labels were translated into english by the authors to ease the readability of the paper.
 The later part of the labels focuses on the underlying pathophysiology  of the discomfort.

Our task is to build a system that makes high quality diagnostic predictions given a discomfort drawing. 
This could be extended into a  decision support system, which could increase the effectiveness and precision of the care for a large group of less favored patients [{\color{blue} \cite{upshur2010they}}]. 

%% file: model.tex
\vspace{-5pt}
\section{Model}
For this application, we adapt IBTM, which is a generative model. 
One advantage of  generative models is that they achieve good performance even on small data sets. 
As it is expensive to collect data in the health care system and there is a big variance in the frequency of different types of diseases, this is highly important. Secondly, generative models have the advantage of being able to handle missing data. 
In this preliminary work, we are only dealing with two modalities, discomfort drawings and diagnostic labels. 
Even in this simplistic setting, the diagnostic labels are not complete. 
In health care systems, there exists a variety of  examinations and tests that are only partially used for different patients.
Hence, a system that can handle missing data is desired in such application. 
Finally, a probabilistic interpretation of the symptoms and diagnostic decisions is desirable. 
IBTM is a factorized multi-modal topic model which enjoys all the properties of generative models and is robust to noise in the data.
  
  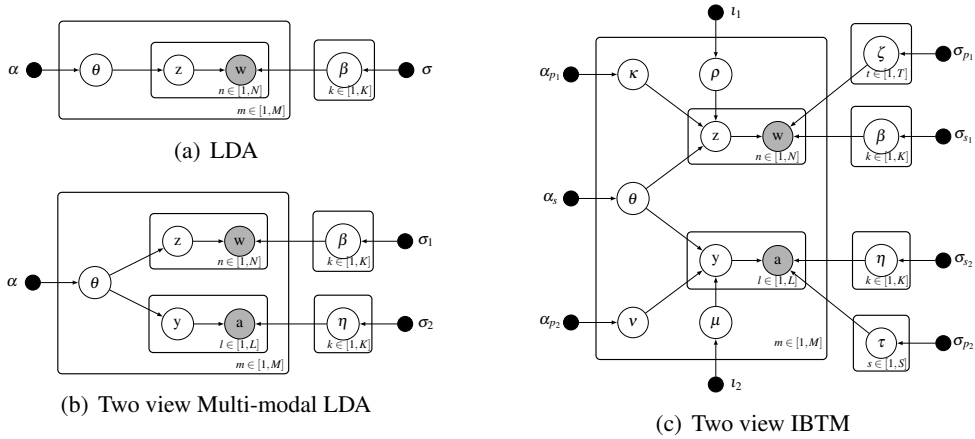
\begin{figure}[h]
\begin{minipage}{.47\textwidth}
\centering
\subfigure[LDA]{
\scalebox{0.55}{\input{tikz/ldaPGM.tex}}
}
\subfigure[Two view Multi-modal LDA]{
\scalebox{0.55}{\input{tikz/PGM_2View.tex}}
}
\end{minipage}~~~
\begin{minipage}{.47\textwidth}
\subfigure[Two view  IBTM]{
\scalebox{0.55}{\input{tikz/PGM.tex}}
}
\end{minipage}%
\vspace{-3mm}
\caption{\footnotesize Graphical representations of three topic models.The nodes with grey shadows indicate an observation while all other nodes are latent variables that need to be learned.}
\label{fig:FMLDA}
\end{figure}
\vspace{-5pt}
\subsection{ Inter-Battery Topic Model} 
\vspace{-3pt}
\paragraph{LDA} Topic models, a group of generative models based on Latent Dirichlet Allocation (LDA) [{\color{blue}\cite{blei03latent}}], have been successfully applied to many scenarios, mainly focusing on discrete data. 
The graphical representation of LDA is shown in Figure~\ref{fig:FMLDA}(a). 
LDA assumes that each word in a document is generated by sampling from a per document topic distribution $\theta \sim Dir(\alpha)$ and per topic word distribution $\beta \sim Dir(\sigma)$. 
The document here stands for an information piece, such as a visual document (picture or video) or any collection of text. 
The topics are latent representations which can be topics in text documents or symptom groups in medical documents. 
\vspace{-3pt}
\paragraph{Multi-Modal Topic Model } LDA is designed for data with a single modality. 
In our application, we want to  jointly model discomfort drawings and the diagnostic labels. 
Additionally, we want to learn a system that is able to give high quality predictions of diagnostic labels given an unseen discomfort drawing. Hence, a multi-modal topic model is needed. 
There exist a number of approaches to extend LDA to capture multi-modal data in a joint fashion [{\color{blue}\cite{blei2003modeling,wang2009simultaneous,wang2011max,hospedales11,zhang13contextual}}], among which some were designed for special applications and based on more assumptions. 
Multi-modal LDA (MMLDA) [{\color{blue}\cite{blei2003modeling,zhang13contextual}}] is the most natural multi-modal  extension of LDA. 
Figure \ref{fig:FMLDA}(b) shows the graphic representation of MMLDA in a two modality case, where $w$ and $a$ represent the observations for each modality and $\theta$ is the joint latent representation.
\vspace{-3pt}
\paragraph{IBTM} As shown in Figure \ref{fig:FMLDA}(b),  MMLDA forces the two modalities to completely share a latent space $\theta$. However,  real-life data is noisy and incomplete in general and might have shared and disjunct latent sources. In our application, we also  need to deal with exchangeable  clinical terms and possible missing labels. IBTM is proposed to make MMLDA more robust with respect to complex real-life data. Compared to MMLDA, a private topic space for each modality ( $\kappa$ and $\nu$ )  is introduced to explain away irrelevant information. By this, the  shared topic space can provide qualitatively better latent representations of the structure of the data.  In IBTM, $\rho \sim Beta(\iota_{1})$ and $\mu \sim Beta(\iota_{2})$ are portions of the information that can be shared between the two modalities, where $\iota_{1}$ and $\iota_{2}$ are two dimensional pairs of beta distribution hyper-parameters.

In our task, the first modality is the discomfort drawing, where a bag-of-words representation of the discomfort areas is used as the observation $w$. 
The second modality is the diagnostic labels $y$ which are only available in the training phase. 
Each document $m$ contains a discomfort drawing and its corresponding diagnostic labels. 
Both modalities share the same per document topic distribution $\theta$ which can be interpreted as the combination of symptoms that generate a drawing and its diagnostic labels. 
For each modality, the private topic distributions $\kappa$ and $\nu$ are used to encode the information that cannot be simultaneously explained by both modalities which are noises per se  in general.
The $\beta$ is the per shared topic distribution for the  drawing locations and $\eta$ is the per  shared topic distribution for the diagnostic labels. These encode the essential information  that will be used for prediction.  Similarly, the $\zeta$ is the per private topic  distribution for  the drawing location and $\tau$ is the per private topic  distribution for the diagnostic labels. These encode irrelevant information that needs to be explained away. Each topic is a latent variable, which indicates the problem of the patient. 
For example, in the case shown in Table \ref{fig:dataExp}, one topic in $\theta$ may be an injury between the 4th and 5th lumber vertebrae, which generates the discomfort drawing ($w$) in the upper tie and lower back and generates the diagnostic labels ($y$) L4 Radiculopathy and DLI L4-L5. 

To learn all  latent parameters in IBTM, mean field variational inference is used in this work, because variational inference is efficient and can easily be adapted to online settings [{\color{blue}\cite{hoffman2010online,ranganath2013adaptive,wang2011online}}]. 
In real health applications, online learning is desirable.  A standard batch update is used for the experiments due to the small amount of the data. An online version of the IBTM for diagnosis  prediction is derived based on the batch vision in \cite{zhang16IBTM} and implemented for long-term usage for this application. 
\vspace{-5pt}
\subsection{Diagnostic Prediction using IBTM} 
\label{sec:DPIBTM}
\vspace{-3pt}
\paragraph{Diagnostic Prediction}
In the training phase, all latent variables will be learned. 
In the testing phase, given an unseen observation $w$ without diagnostic labels $y$, we will estimate the per document distributions $\theta$ and $\kappa$ using the learned per topic word distribution i.e.~ the global parameters $\beta$ and $\zeta$.  Given the estimated $\theta$ for the new document, possible $y$ can be easily generated with the help of the learned parameter $\eta$. Hence, we can predict diagnostic labels given a new discomfort drawing. Using IBTM, only the shared topic distribution $\theta$ is used for diagnostic prediction, which is similar to MMLDA. However, the latent representation is of higher quality due to the private topics $\kappa$ that explain away irrelevant information. 

\begin{figure}[h]
\centering
\subfigure[16 Clusters]{
\includegraphics[height=5cm]{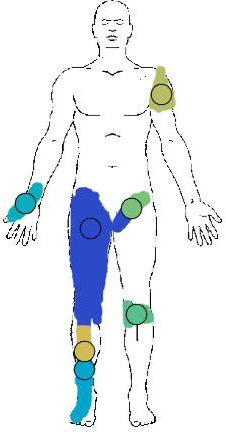}
\includegraphics[height=5cm]{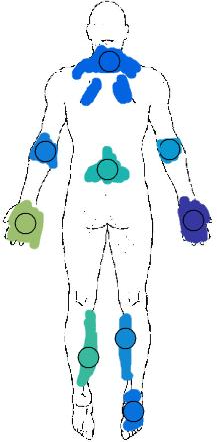}
}
~~~~~~~~
\subfigure[7 Clusters]{
\includegraphics[height=5cm]{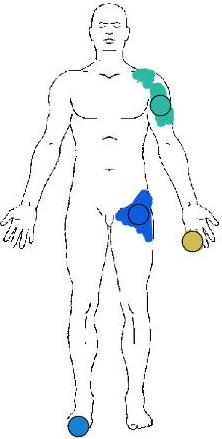}
\includegraphics[height=5cm]{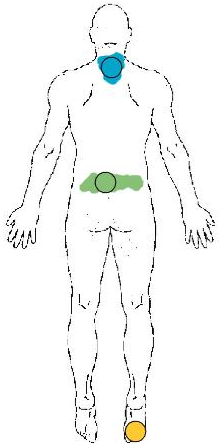}
}
\caption{\footnotesize Examples of the clustering output with mean shift. The shades were drawn by patients while the circles indicate the mean location of an identified cluster of the discomfort region. }
\label{fig:meanshift}
\vspace{-1cm}
\end{figure}

Using IBTM, we can generate all possible diagnostic labels $y$ for a drawing with different probabilities. However, it is difficult to decide how many diagnostic labels are actually  needed since there exists no universal probability threshold. One patient may have a broken toe for which one or two labels are needed. Another patient may have several discoligament injuries causing discomfort in multiple areas, where more than 50 labels may be needed. To determine how many diagnostic labels are required, we would like to know how many discomfort regions are contained in the test pain drawing. Intuitively, the number of diagnostic labels are positively correlated with the number of discomfort regions. Thus, we use the mean shift clustering algorithm [{\color{blue}\cite{comaniciu2002mean}}] to cluster the  noisy, irregular drawing locations into coherent groups. Mean shift clustering is non-parametric, hence it let data determine  the number of clusters. 
Figure \ref{fig:meanshift} shows examples of the output of mean shift clustering on test discomfort drawings. 
In this paper, we use twice the number of clusters as the number of prediction labels since the labels are a mixture of symptom diagnostic labels and pattern/pathophysiological diagnostic  labels.  

\vspace{-3pt}
\paragraph{Diagnostic Interpretation} Besides automatic diagnostic prediction,  it is useful to investigate which features the models learn to make these diagnostic predictions. Based on big datasets, the model may give us insights into diagnostic procedures. Hence, in this work, we also  generate synthetic discomfort drawings given a diagnostic label. In an ideal case, the model will provide  knowledge about the typical discomfort drawing of each diagnostic  label. This is done in a similar way as diagnostic prediction. Instead of a test drawing, we give the model a test diagnostic label without a drawing. In the following, the model is used to generate  possible drawings. In the experiments section, we will evaluate the diagnostic interpretation of these generated synthetic discomfort drawings.

%% file: tikz/ldaPGM.tex
\pgfdeclarelayer{background}
\pgfdeclarelayer{foreground}
\pgfsetlayers{background,main,foreground}

\begin{tikzpicture}

\tikzstyle{surround} = [thick,draw=black,rounded corners=1mm]

\tikzstyle{scalarnode} = [circle, draw, fill=white!11,  
    text width=1.2em, text badly centered, inner sep=2.5pt]

\tikzstyle{scalarnodeCyan} = [circle, draw=cyan, fill=white!11,  
    text width=1.2em, text badly centered, inner sep=2.5pt]
\tikzstyle{discnode}=[rectangle,draw,fill=white!11,minimum size=0.9cm]

\tikzstyle{Vnode}=[circle, radius=1pt,draw,fill=black]
\tikzstyle{vectornode} = [circle, draw, fill=white!11,  
    text width=2.3em, text badly centered, inner sep=2pt]
\tikzstyle{state} = [rectangle, draw, text centered, fill=white, 
    text width=8em, text height=6.7em, rounded corners]

\tikzstyle{arrowline} = [draw,color=black, -latex]
\tikzstyle{carrowline} = [line width=2pt, draw,color=black, -latex]
\tikzstyle{line} = [draw]

\node [Vnode] at ( 0.5, 1.5) (alpha_s){};
\node [] at ( 0, 1.5) (){$\alpha$};
\node [scalarnode] at ( 2, 1.5 ) (theta) { $\theta$ };
\node [scalarnode] at ( 4, 1.5) (z) {z};
\node [scalarnode, fill=black!30] at ( 5.5, 1.5) (w) {w};
\node [scalarnode] at ( 8, 1.5) (beta) {$\beta$};
\node [Vnode] at ( 9.5, 1.5) (sigma) {};
\node [] at ( 10, 1.5) () {$\sigma$};

\node[surround, inner sep = .3cm] (f_N) [fit = (z)(w) ] {};
\node[surround, inner sep = .5cm] (f_M) [fit = (f_N)(theta)] {};
\node[surround, inner sep = .3cm] (f_beta) [fit = (beta)] {};

\node [] at (6, 0.5) (M) {\scriptsize $m \in [1,M]$};
\node [] at (5.5, 1) (N) {\scriptsize $n \in [1,N]$};
\node [] at (8.15, 1) () {\scriptsize $k \in [1,K]$};

\path [arrowline] (alpha_s) to (theta); 
\path [arrowline] (theta) to (z); 
\path [arrowline] (z) to (w); 
\path [arrowline] (beta) to (w); 
\path [arrowline] (sigma) to (beta); 

\end{tikzpicture}

%% file: tikz/PGM_2View.tex
\pgfdeclarelayer{background}
\pgfdeclarelayer{foreground}
\pgfsetlayers{background,main,foreground}

\begin{tikzpicture}

\tikzstyle{surround} = [thick,draw=black,rounded corners=1mm]

\tikzstyle{scalarnode} = [circle, draw, fill=white!11,  
    text width=1.2em, text badly centered, inner sep=2.5pt]

\tikzstyle{scalarnodeCyan} = [circle, draw=cyan, fill=white!11,  
    text width=1.2em, text badly centered, inner sep=2.5pt]
\tikzstyle{discnode}=[rectangle,draw,fill=white!11,minimum size=0.9cm]

\tikzstyle{Vnode}=[circle, radius=1pt,draw,fill=black]
\tikzstyle{vectornode} = [circle, draw, fill=white!11,  
    text width=2.3em, text badly centered, inner sep=2pt]
\tikzstyle{state} = [rectangle, draw, text centered, fill=white, 
    text width=8em, text height=6.7em, rounded corners]

\tikzstyle{arrowline} = [draw,color=black, -latex]
\tikzstyle{carrowline} = [line width=2pt, draw,color=black, -latex]
\tikzstyle{line} = [draw]

\node [Vnode] at ( 0.5, 0) (alpha_s){};
\node [] at ( 0, 0) (){$\alpha$};
\node [scalarnode] at ( 2, 0 ) (theta) { $\theta$ };

\node [scalarnode] at ( 4, 1) (z) {z};
\node [scalarnode, fill=black!30] at ( 5.5, 1) (w) {w};

\node [scalarnode] at ( 4, -1) (y) {y};
\node [scalarnode, fill=black!30] at ( 5.5, -1) (a) {a};

\node [scalarnode] at ( 8, 1) (beta) {$\beta$};
\node [Vnode] at ( 9.5, 1) (sigma_s1) {};
\node [] at ( 10, 1) () {$\sigma_{1}$};

\node [scalarnode] at ( 8, -1) (eta) {$\eta$};
\node [Vnode] at ( 9.5, -1) (sigma_s2) {};
\node [] at ( 10, -1) () {$\sigma_{2}$};

\node[surround, inner sep = .3cm] (f_N) [fit = (z)(w) ] {};
\node[surround, inner sep = .3cm] (f_L) [fit = (y)(a) ] {};
\node[surround, inner sep = .5cm] (f_M) [fit = (f_N)(f_L)(theta) ] {};
\node[surround, inner sep = .3cm] (f_beta) [fit = (beta) ] {};
\node[surround, inner sep = .3cm] (f_eta) [fit = (eta) ] {};

\node [] at (6, -2) (M) {\scriptsize $m \in [1,M]$};
\node [] at (5.5, -2+0.5) (L) {\scriptsize $l \in [1,L]$};
\node [] at (5.5, 1-0.5) (N) {\scriptsize $n \in [1,N]$};;
\node [] at (8.15, 0.5) () {\scriptsize $k \in [1,K]$};
\node [] at (8.15, -1.5) () {\scriptsize $k \in [1,K]$};

\path [arrowline] (alpha_s) to (theta); 
\path [arrowline] (theta) to (z); 
\path [arrowline] (z) to (w);

\path [arrowline] (theta) to (y); 
\path [arrowline] (y) to (a);

\path [arrowline] (beta) to (w); 
\path [arrowline] (sigma_s1) to (beta); 
\path [arrowline] (eta) to (a); 
\path [arrowline] (sigma_s2) to (eta); 
\end{tikzpicture}

%% file: tikz/PGM.tex
\pgfdeclarelayer{background}
\pgfdeclarelayer{foreground}
\pgfsetlayers{background,main,foreground}

\begin{tikzpicture}

\tikzstyle{surround} = [thick,draw=black,rounded corners=1mm]

\tikzstyle{scalarnode} = [circle, draw, fill=white!11,  
    text width=1.2em, text badly centered, inner sep=2.5pt]

\tikzstyle{scalarnodeCyan} = [circle, draw=cyan, fill=white!11,  
    text width=1.2em, text badly centered, inner sep=2.5pt]
\tikzstyle{discnode}=[rectangle,draw,fill=white!11,minimum size=0.9cm]

\tikzstyle{Vnode}=[circle, radius=1pt,draw,fill=black]
\tikzstyle{vectornode} = [circle, draw, fill=white!11,  
    text width=2.3em, text badly centered, inner sep=2pt]
\tikzstyle{state} = [rectangle, draw, text centered, fill=white, 
    text width=8em, text height=6.7em, rounded corners]

\tikzstyle{arrowline} = [draw,color=black, -latex]
\tikzstyle{carrowline} = [line width=2pt, draw,color=black, -latex]
\tikzstyle{line} = [draw]

\node [Vnode] at ( 0.5, 0) (alpha_s){};
\node [] at ( 0, 0) (){$\alpha_s$};
\node [scalarnode] at ( 2, 0 ) (theta) { $\theta$ };
\node [Vnode] at ( 0.5, 3) (alpha_p1){};
\node [] at ( 0, 3) (){$\alpha_{p_1}$};
\node [scalarnode] at ( 2 , 3 ) (kappa) { $\kappa$ };
\node [scalarnode] at ( 4, 1.5) (z) {z};
\node [scalarnode, fill=black!30] at ( 5.5, 1.5) (w) {w};

\node [scalarnode] at ( 4, 3) (rho) {$\rho$};
\node [Vnode] at ( 4, 4.5) (iota1) {};
\node [] at ( 4.5, 4.5) () {$\iota_1$};

\node [Vnode] at ( 0.5, -3) (alpha_p2){};
\node [] at ( 0, -3) (){$\alpha_{p_2}$};
\node [scalarnode] at ( 2 , -3 ) (nu) { $\nu$ };
\node [scalarnode] at ( 4, -1.5) (y) {y};
\node [scalarnode, fill=black!30] at ( 5.5, -1.5) (a) {a};

\node [scalarnode] at ( 4, -3) (mu) {$\mu$};
\node [Vnode] at ( 4, -4.5) (iota2) {};
\node [] at ( 4.5, -4.5) () {$\iota_2$};

\node [scalarnode] at ( 8, 1.5+2) (zeta) {$\zeta$};
\node [Vnode] at ( 9.5, 1.5+2) (sigma_p1) {};
\node [] at ( 10, 1.5+2) () {$\sigma_{p_1}$};

\node [scalarnode] at ( 8, 1.5) (beta) {$\beta$};
\node [Vnode] at ( 9.5, 1.5) (sigma_s1) {};
\node [] at ( 10, 1.5) () {$\sigma_{s_1}$};

\node [scalarnode] at ( 8, -1.5) (eta) {$\eta$};
\node [Vnode] at ( 9.5, -1.5) (sigma_s2) {};
\node [] at ( 10, -1.5) () {$\sigma_{s_2}$};

\node [scalarnode] at ( 8, -1.5-2) (tau) {$\tau$};
\node [Vnode] at ( 9.5, -1.5-2) (sigma_p2) {};
\node [] at ( 10, -1.5-2) () {$\sigma_{p_2}$};
\node[surround, inner sep = .3cm] (f_N) [fit = (z)(w) ] {};
\node[surround, inner sep = .3cm] (f_L) [fit = (y)(a) ] {};
\node[surround, inner sep = .5cm] (f_M) [fit = (f_N)(f_L)(theta)(rho)(mu) ] {};

\node[surround, inner sep = .3cm] (f_zeta) [fit = (zeta) ] {};
\node[surround, inner sep = .3cm] (f_beta) [fit = (beta) ] {};
\node[surround, inner sep = .3cm] (f_eta) [fit = (eta) ] {};
\node[surround, inner sep = .3cm] (f_tau) [fit = (tau) ] {};

\node [] at (6, -3.5) (M) {\scriptsize $m \in [1,M]$};
\node [] at (5.5, -2) (L) {\scriptsize $l \in [1,L]$};
\node [] at (5.5, 1) (N) {\scriptsize $n \in [1,N]$};
\node [] at (8.15, 3) () {\scriptsize $t \in [1,T]$};
\node [] at (8.15, 1) () {\scriptsize $k \in [1,K]$};
\node [] at (8.15, -2) () {\scriptsize $k \in [1,K]$};
\node [] at (8.15, -4) () {\scriptsize $s \in [1,S]$};

\path [arrowline] (alpha_s) to (theta); 
\path [arrowline] (alpha_p1) to (kappa); 
\path [arrowline] (theta) to (z); 
\path [arrowline] (kappa) to (z); 
\path [arrowline] (rho) to (z); 
\path [arrowline] (z) to (w); 
\path [arrowline] (iota1) to (rho);

\path [arrowline] (alpha_p2) to (nu); 
\path [arrowline] (theta) to (y); 
\path [arrowline] (nu) to (y); 
\path [arrowline] (mu) to (y); 
\path [arrowline] (y) to (a); 
\path [arrowline] (iota2) to (mu); 

\path [arrowline] (zeta) to (w); 
\path [arrowline] (sigma_p1) to (zeta); 
\path [arrowline] (beta) to (w); 
\path [arrowline] (sigma_s1) to (beta); 
\path [arrowline] (eta) to (a); 
\path [arrowline] (sigma_s2) to (eta); 
\path [arrowline] (tau) to (a); 
\path [arrowline] (sigma_p2) to (tau); 
\end{tikzpicture}

%% file: exp.tex
\vspace{-5pt}  
\section{Experiments}
\vspace{-3pt}
\paragraph{Dataset} A dataset of 174 real-world patient discomfort drawings was collected from clinical records with diagnostic labels from medical experts.
The clinic in question is specialized on diagnosing unspecific pain and discomfort and presented patient cases often have neuropathic pain syndromes.
Since bilateral diagnostic labels indicates the problem shows in both sides, we preprocess the data breaking all bilateral labels into left side and right side labels. Taken the example in Figure \ref{fig:dataExp}, the preprocessed labels are:
\\
{\footnotesize Interscapular discomfort; R arm discomfort; L hand discomfort; R hand discomfort; Lumbago; L crest of the ilium discomfort; R crest of the ilium discomfort; L side thigh discomfort; L back thigh discomfort; R back thigh discomfort; L calf discomfort; R calf discomfort; L achilles tendinitis; R achilles tendinitis; L  shin discomfort; R  shin discomfort; R  inguinal discomfort;	L L5 Radiculopathy; R L5 Radiculopathy; L S1 Radiculopathy; R S1 Radiculopathy; L C7 Radiculopathy; R C7 Radiculopathy; DLI L4-L5; DLI S1-S2; DLI C6-C7.}
\\
Symptoms such as  Interscapular discomfort are kept using only the name without the indication of left or right since it lies in the middle of the body. However, when the discomfort is occurring on one side of the body, medical experts also indicate this information. Moreover, it is common that the same symptoms can be termed differently in different systems. After consulting medical personal, we treat the diagnostic labels listed in Table \ref{tab:Exchange} as exchangeable, which means that they will be treated as the same label. Whether these medical terms are equivalent is a point of discussion, but this is not in the range of this work. We believe that the equivalence of the labels listed in Table \ref{tab:Exchange} is assured.
After preprocessing the diagnostic labels, the number of labels per patient ranged from 2 to 50.  Figure~\ref{fig:hist} shows a histogram of the top symptom,  pattern and pathophysiological diagnostic labels. About $30\%$ of these diagnostic labels appear only once in the dataset.

\begin{table}[h]
\centering
\begin{tabular}{ | c  | c || c | c|}
\hline
\multicolumn{2}{|c||}{Exchangeable labels} &\multicolumn{2}{|c|}{Exchangeable labels}\\
\hline
Medial elbow dcf&	Golfer's elbow&
Lateral elbow dcf &	Tennis elbow\\
Nerve strain effect &	Myelopathy &
Medial knee arthrosis  &	Gonarthrosis\\
Medial meniscus	&Medial gonarthrosis &
Jew dcf &	Bruxism \\
Back thigh dcf &	Hamstrings dcf  &
Hand joint dcf & Carpal Tunnel Syndrome\\
Heel dcf &	 Calcaneodynia &
Upper abdominal dcf	& Gastritis\\
Side thigh dcf & Piriformis tendonitis&
Crest of the ilium dcf & Trochanter \\
Throat dcf	& Globus hystericus &
Coxarthrosis &  	Hip joint arthritis \\
\hline
\end{tabular}
\caption{\footnotesize List of exchangeable labels. "dcf" stands for discomfort.}
\label{tab:Exchange}
\end{table}
\vspace{-5pt}

\begin{figure}[h]
\centering
\subfigure[Symptom diagnostic labels ]{
\scalebox{0.9}{
\begin{tikzpicture}
    \node[] at (0,0){\includegraphics[trim=0 100 0 0, clip ,width =13.5cm]{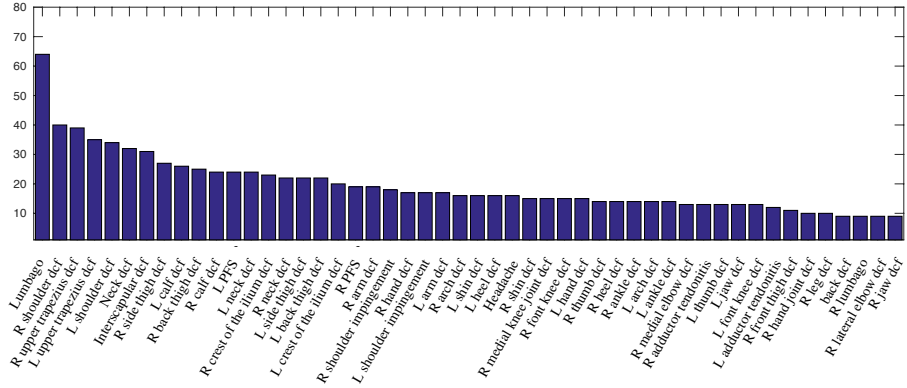} };
    \node[rotate=60,align=flush right, text width=2.21cm] at (-6.723,-3.003){\tiny Lumbago}; 
    \node[rotate=60,align=flush right, text width=2.21cm] at (-6.723+0.259,-3.003){\tiny R shoulder dcf}; 
     \node[rotate=60,align=flush right, text width=2.21cm] at (-6.723+2*0.259,-3.003){\tiny R upper trapezius dcf}; 
      \node[rotate=60,align=flush right, text width=2.21cm] at (-6.723+3*0.259,-3.003){\tiny L upper trapezius dcf}; 
      \node[rotate=60,align=flush right, text width=2.21cm] at (-6.723+4*0.259,-3.003){\tiny L shoulder dcf};
     \node[rotate=60,align=flush right, text width=2.21cm] at (-6.723+5*0.259,-3.003){\tiny Neck dcf };
     \node[rotate=60,align=flush right, text width=2.21cm] at (-6.723+6*0.259,-3.003){\tiny Interscapular dcf};
      \node[rotate=60,align=flush right, text width=2.21cm] at (-6.723+7*0.259,-3.003){\tiny R side thigh dcf};
     \node[rotate=60,align=flush right, text width=2.21cm] at (-6.723+8*0.259,-3.003){\tiny L calf dcf};
     \node[rotate=60,align=flush right, text width=2.21cm] at (-6.723+9*0.259,-3.003){\tiny R back thigh dcf};
      \node[rotate=60,align=flush right, text width=2.21cm] at (-6.723+10*0.259,-3.003){\tiny R calf dcf};
     \node[rotate=60,align=flush right, text width=2.21cm] at (-6.723+11*0.259,-3.003){\tiny L PFS};
     \node[rotate=60,align=flush right, text width=2.21cm] at (-6.723+12*0.259,-3.003){\tiny L neck dcf};
      \node[rotate=60,align=flush right, text width=2.21cm] at (-6.723+13*0.259,-3.003){\tiny R crest of the ilium dcf};
           \node[rotate=60,align=flush right, text width=2.21cm] at (-6.723+14*0.259,-3.003){\tiny R neck dcf};
     \node[rotate=60,align=flush right, text width=2.21cm] at (-6.723+15*0.259,-3.003){\tiny L side thigh dcf };
      \node[rotate=60,align=flush right, text width=2.21cm] at (-6.723+16*0.259,-3.003){\tiny L back thigh dcf};
           \node[rotate=60,align=flush right, text width=2.21cm] at (-6.723+17*0.259,-3.003){\tiny L crest of the ilium dcf};
     \node[rotate=60,align=flush right, text width=2.21cm] at (-6.723+18*0.259,-3.003){\tiny R PFS};
      \node[rotate=60,align=flush right, text width=2.21cm] at (-6.723+19*0.259,-3.003){\tiny R arm dcf};
           \node[rotate=60,align=flush right, text width=2.21cm] at (-6.723+20*0.259,-3.003){\tiny R shoulder impingement};
     \node[rotate=60,align=flush right, text width=2.21cm] at (-6.723+21*0.259,-3.003){\tiny R hand dcf};
      \node[rotate=60,align=flush right, text width=2.21cm] at (-6.723+22*0.259,-3.003){\tiny L shoulder impingement};
      \node[rotate=60,align=flush right, text width=2.21cm] at (-6.723+23*0.259,-3.003){\tiny L arm dcf};
       \node[rotate=60,align=flush right, text width=2.21cm] at (-6.723+24*0.259,-3.003){\tiny R arch dcf};
      \node[rotate=60,align=flush right, text width=2.21cm] at (-6.723+25*0.259,-3.003){\tiny L shin dcf};
       \node[rotate=60,align=flush right, text width=2.21cm] at (-6.723+26*0.259,-3.003){\tiny L heel dcf};
       \node[rotate=60,align=flush right, text width=2.21cm] at (-6.723+27*0.259,-3.003){\tiny Headache};
          \node[rotate=60,align=flush right, text width=2.21cm] at (-6.723+28*0.259,-3.003){\tiny R shin dcf};
       \node[rotate=60,align=flush right, text width=2.21cm] at (-6.723+29*0.259,-3.003){\tiny R medial knee joint dcf};
      \node[rotate=60,align=flush right, text width=2.21cm] at (-6.723+30*0.259,-3.003){\tiny R font knee dcf};
       \node[rotate=60,align=flush right, text width=2.21cm] at (-6.723+31*0.259,-3.003){\tiny L hand dcf};
      \node[rotate=60,align=flush right, text width=2.21cm] at (-6.723+32*0.259,-3.003){\tiny R thumb dcf};
       \node[rotate=60,align=flush right, text width=2.21cm] at (-6.723+33*0.259,-3.003){\tiny R heel dcf};
      \node[rotate=60,align=flush right, text width=2.21cm] at (-6.723+34*0.259,-3.003){\tiny R ankle dcf};
       \node[rotate=60,align=flush right, text width=2.21cm] at (-6.723+35*0.259,-3.003){\tiny L arch dcf};
      \node[rotate=60,align=flush right, text width=2.21cm] at (-6.723+36*0.259,-3.003){\tiny L ankle dcf};
       \node[rotate=60,align=flush right, text width=2.21cm] at (-6.723+37*0.259,-3.003){\tiny R medial elbow dcf};
          \node[rotate=60,align=flush right, text width=2.21cm] at (-6.723+38*0.259,-3.003){\tiny R adductor tendonitis};
           \node[rotate=60,align=flush right, text width=2.21cm] at (-6.723+39*0.259,-3.003){\tiny L thumb dcf};
     \node[rotate=60,align=flush right, text width=2.21cm] at (-6.723+40*0.259,-3.003){\tiny L jaw dcf};
           \node[rotate=60,align=flush right, text width=2.21cm] at (-6.723+41*0.259,-3.003){\tiny L font knee dcf};
      \node[rotate=60,align=flush right, text width=2.21cm] at (-6.723+42*0.259,-3.003){\tiny L adductor tendonitis};
       \node[rotate=60,align=flush right, text width=2.21cm] at (-6.723+43*0.259,-3.003){\tiny R front thigh dcf};
      \node[rotate=60,align=flush right, text width=2.21cm] at (-6.723+44*0.259,-3.003){\tiny R hand joint dcf};
       \node[rotate=60,align=flush right, text width=2.21cm] at (-6.723+45*0.259,-3.003){\tiny R leg dcf};
      \node[rotate=60,align=flush right, text width=2.21cm] at (-6.723+46*0.259,-3.003){\tiny back dcf};
       \node[rotate=60,align=flush right, text width=2.21cm] at (-6.723+47*0.259,-3.003){\tiny R lumbago};
      \node[rotate=60,align=flush right, text width=2.21cm] at (-6.723+48*0.259,-3.003){\tiny R lateral elbow dcf};
       \node[rotate=60,align=flush right, text width=2.21cm] at (-6.723+49*0.259,-3.003){\tiny R jaw dcf};
     \end{tikzpicture}   
     }  
    }
    

\subfigure[Pattern diagnostic labels ]{
\scalebox{0.9}{
\begin{tikzpicture}
    \node[] at (0,0){\includegraphics[trim=0 70 0 0, clip ,width =6.5cm]{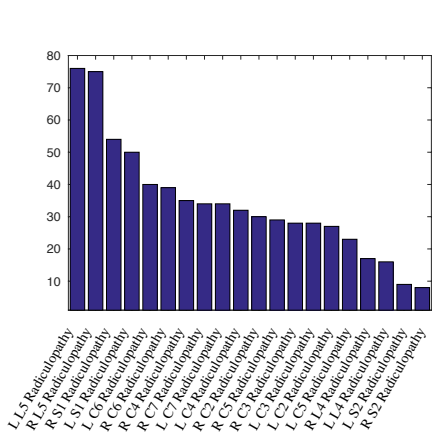} };
    \node[rotate=60,align=flush right, text width=1.9cm] at (-2.923+0.274,-3.003){\tiny L L5 Radiculopathy}; 
    \node[rotate=60,align=flush right, text width=1.9cm] at (-2.923+2*0.274,-3.003){\tiny R L5 Radiculopathy}; 
      \node[rotate=60,align=flush right, text width=1.9cm] at (-2.923+3*0.274,-3.003){\tiny R S1 Radiculopathy};  
       \node[rotate=60,align=flush right, text width=1.9cm] at (-2.923+4*0.274,-3.003){\tiny L S1  Radiculopathy}; 
        \node[rotate=60,align=flush right, text width=1.9cm] at (-2.923+5*0.274,-3.003){\tiny L C6  Radiculopathy}; 
         \node[rotate=60,align=flush right, text width=1.9cm] at (-2.923+6*0.274,-3.003){\tiny R C6 Radiculopathy};  
          \node[rotate=60,align=flush right, text width=1.9cm] at (-2.923+7*0.274,-3.003){\tiny R C4 Radiculopathy};  
          \node[rotate=60,align=flush right, text width=1.9cm] at (-2.923+8*0.274,-3.003){\tiny R C7 Radiculopathy}; 
          \node[rotate=60,align=flush right, text width=1.9cm] at (-2.923+9*0.274,-3.003){\tiny L C7 Radiculopathy};  
          \node[rotate=60,align=flush right, text width=1.9cm] at (-2.923+10*0.274,-3.003){\tiny L C4 Radiculopathy};  
          \node[rotate=60,align=flush right, text width=1.9cm] at (-2.923+11*0.274,-3.003){\tiny R C2 Radiculopathy};  
          \node[rotate=60,align=flush right, text width=1.9cm] at (-2.923+12*0.274,-3.003){\tiny R C5 Radiculopathy}; 
          \node[rotate=60,align=flush right, text width=1.9cm] at (-2.923+13*0.274,-3.003){\tiny R C3  Radiculopathy}; 
          \node[rotate=60,align=flush right, text width=1.9cm] at (-2.923+14*0.274,-3.003){\tiny L C3  Radiculopathy}; 
          \node[rotate=60,align=flush right, text width=1.9cm] at (-2.923+15*0.274,-3.003){\tiny L C2 Radiculopathy};  
          \node[rotate=60,align=flush right, text width=1.9cm] at (-2.923+16*0.274,-3.003){\tiny L C5 Radiculopathy};  
          \node[rotate=60,align=flush right, text width=1.9cm] at (-2.923+17*0.274,-3.003){\tiny R L4 Radiculopathy};  
          \node[rotate=60,align=flush right, text width=1.9cm] at (-2.923+18*0.274,-3.003){\tiny L L4 Radiculopathy}; 
          \node[rotate=60,align=flush right, text width=1.9cm] at (-2.923+19*0.274,-3.003){\tiny L S2  Radiculopathy}; 
          \node[rotate=60,align=flush right, text width=1.9cm] at (-2.923+20*0.274,-3.003){\tiny R S2   Radiculopathy};          
         \end{tikzpicture}   
     }  
    }
    \subfigure[Pathophysiological diagnostic labels ]{
\scalebox{0.9}{
\begin{tikzpicture}
    \node[] at (0,0){\includegraphics[trim=0 93 0 0, clip ,width =6.5cm]{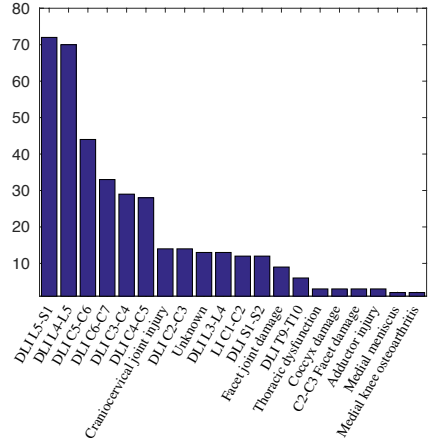} };
    \node[rotate=60,align=flush right, text width=2.35cm] at (-3.323+0.284,-3.303){\tiny DLI L5-S1}; 
    \node[rotate=60,align=flush right, text width=2.35cm] at (-3.323+2*0.284,-3.303){\tiny DLI L4-L5}; 
    \node[rotate=60,align=flush right, text width=2.35cm] at (-3.323+3*0.284,-3.303){\tiny DLI C5-C6}; 
    \node[rotate=60,align=flush right, text width=2.35cm] at (-3.323+4*0.284,-3.303){\tiny DLI C6-C7}; 
    \node[rotate=60,align=flush right, text width=2.35cm] at (-3.323+5*0.284,-3.303){\tiny DLI C3-C4}; 
    \node[rotate=60,align=flush right, text width=2.35cm] at (-3.323+6*0.284,-3.303){\tiny DLI C4-C5}; 
    \node[rotate=60,align=flush right, text width=2.35cm] at (-3.323+7*0.284,-3.303){\tiny Craniocervical joint injury}; 
    \node[rotate=60,align=flush right, text width=2.35cm] at (-3.323+8*0.284,-3.303){\tiny DLI C2-C3}; 
    \node[rotate=60,align=flush right, text width=2.35cm] at (-3.323+9*0.284,-3.303){\tiny Unknown}; 
    \node[rotate=60,align=flush right, text width=2.35cm] at (-3.323+10*0.284,-3.303){\tiny DLI L3-L4}; 
    \node[rotate=60,align=flush right, text width=2.35cm] at (-3.323+11*0.284,-3.303){\tiny LI C1-C2}; 
    \node[rotate=60,align=flush right, text width=2.35cm] at (-3.323+12*0.284,-3.303){\tiny DLI S1-S2}; 
    \node[rotate=60,align=flush right, text width=2.35cm] at (-3.323+13*0.284,-3.303){\tiny Facet joint damage}; 
    \node[rotate=60,align=flush right, text width=2.35cm] at (-3.323+14*0.284,-3.303){\tiny DLI T9-T10}; 
    \node[rotate=60,align=flush right, text width=2.35cm] at (-3.323+15*0.284,-3.303){\tiny Thoracic dysfunction}; 
    \node[rotate=60,align=flush right, text width=2.35cm] at (-3.323+16*0.284,-3.303){\tiny Coccyx damage};
    \node[rotate=60,align=flush right, text width=2.35cm] at (-3.323+17*0.284,-3.303){\tiny C2-C3 Facet damage};
     \node[rotate=60,align=flush right, text width=2.35cm] at (-3.323+18*0.284,-3.303){\tiny Adductor injury};
     \node[rotate=60,align=flush right, text width=2.35cm] at (-3.323+19*0.284,-3.303){\tiny Medial meniscus};
     \node[rotate=60,align=flush right, text width=2.35cm] at (-3.323+20*0.284,-3.303){\tiny Medial knee osteoarthritis};
             \end{tikzpicture}   
     }  
    }
    
\caption{\footnotesize Histogram of different types of diagnostic labels appeared in the dataset. The $x$-axis shows different diagnostic labels and the $y$-axis shows the number of occurrences of each label.}
\vspace{-1cm}
\label{fig:hist}
\end{figure}

\vspace{-5pt}
\subsection{Diagnostic Prediction Evaluation} 
We randomly split the dataset into two halves.
One half is used for training the model where both discomfort drawings and diagnostic labels are used while the other half is used for testing (i.e. only the discomfort drawings are available). 
We cluster all  painted point locations on the drawing using K-means clustering with $256$ clusters. 
Subsequently, each discomfort drawing is represented with help of a bag-of-location words.  
In this work, we only use those discomfort area which have been confirmed by medical experts to be the most relevant. 
The diagnostic labels are used as the second modality. 
To balance the number of words in both modalities [{\color{blue}\cite{tang15understanding,zhang14how}}], the diagnostic labels are scaled up by 10 in the experiment so that the number of words in both modalities is in the same order of magnitude.

We use the average F-measure on the predicted diagnostic terms to evaluate the prediction performance. The number of predicted diagnostic terms is  determined by mean shift clustering for each test drawing. Additionally, we set a minimum number of 5 labels and a maximum number of 50 labels. 
The F-measure is defined as:
\begin{equation}
\text{F-measure}=\frac{2\times \text{Precision} \times \text{Recall}}{\text{Precision} + \text{Recall}}.
\end{equation}

The dataset is randomly split 10 times for evaluation and the performance is reported in Table \ref{tab:PredictionPrf} with mean and standard deviation for these 10 groups of experiments with different numbers of shared topics\footnote{For each experiment setting, 10 random seeds were considered and the best result is used.}. The number of private topics is set to $T=5, S=5$ in all the experiments. The hyper-parameters are set to $\alpha_{*}=0.8$, $\sigma_{*}=0.6$ and $\iota_{*}=(1,1)$.

\begin{table}[h]
\centering
\begin{tabular}{ | c| c | c | c|c | c|}
\hline
& $K=5$ & $K=10$ & $K=20$& $K=30$& $K=50$ \\
\hline
\small F-measure& \small $34.31 \pm 1.35\%$ &\small $36.7 \pm 1.37\%$ &\small $38.32 \pm 1.1\%$ &\small $38.56 \pm 1.23\%$ &\small $38.81 \pm 1.08\%$\\
\hline
\end{tabular}
\caption{\footnotesize Prediction performance}
\label{tab:PredictionPrf}
\end{table}
\vspace{-5pt}

\begin{table}
\centering
\begin{tabular}{ | c| p{10.9cm} |}
\hline
\multirow{4}{*}{\includegraphics[height=3cm]{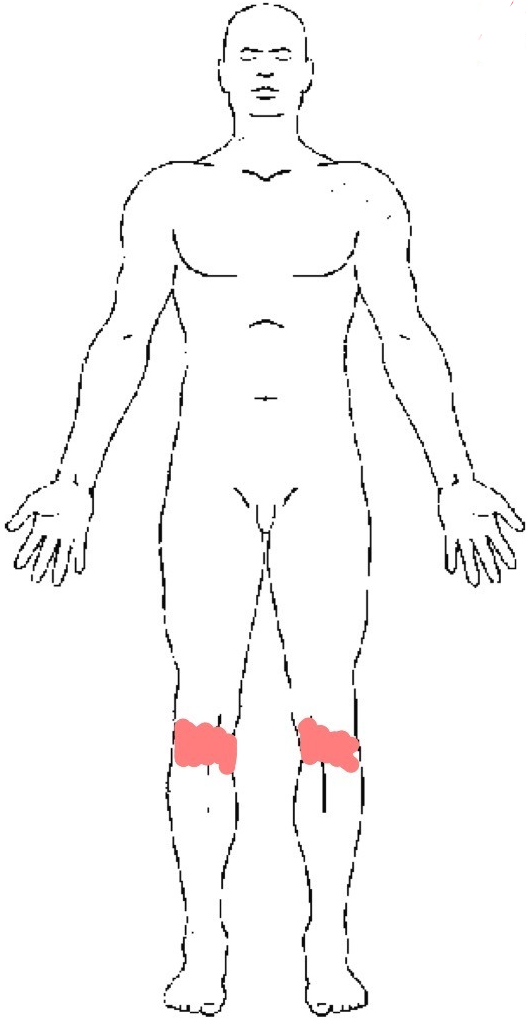}
\includegraphics[height=3cm]{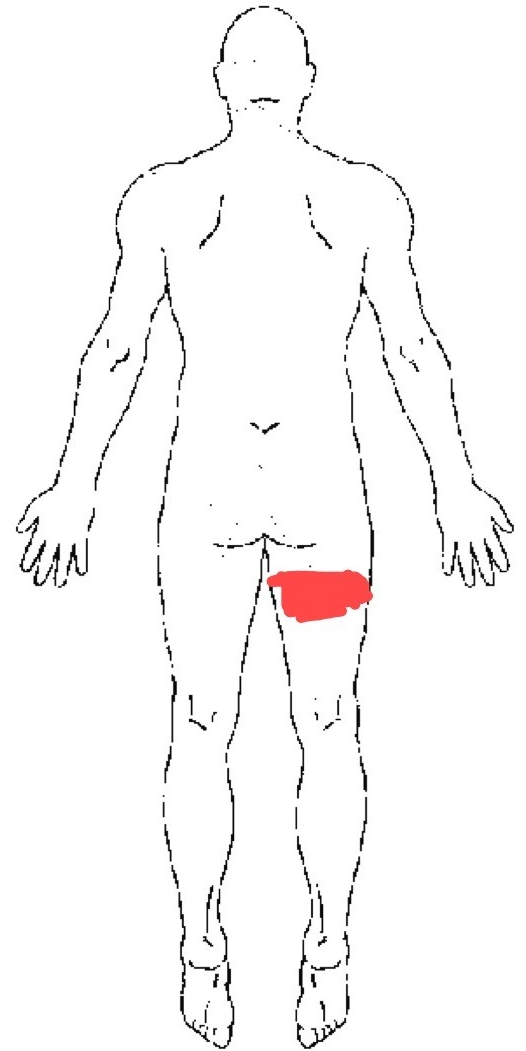}
}
&\scriptsize \textbf{6 Prd:} \color{blue}R back thigh dcf; L PFS (Patellofemoral pain syndrome); R PFS;\\
&\scriptsize \color{blue} 	L L5 Rdc; R L5 Rdc;	DLI L4-L5;\\[5pt]
&\scriptsize \textbf{6 GT:}  \color{blue}R back thigh dcf; L PFS; R PFS;\\
&\scriptsize \color{blue} 	L L5 Rdc; R L5 Rdc;	DLI L4-L5;\\[15pt]
\hline
\raisebox{-1.5cm}{\multirow{3}{*}{ \vspace{10pt}\includegraphics[height=3cm]{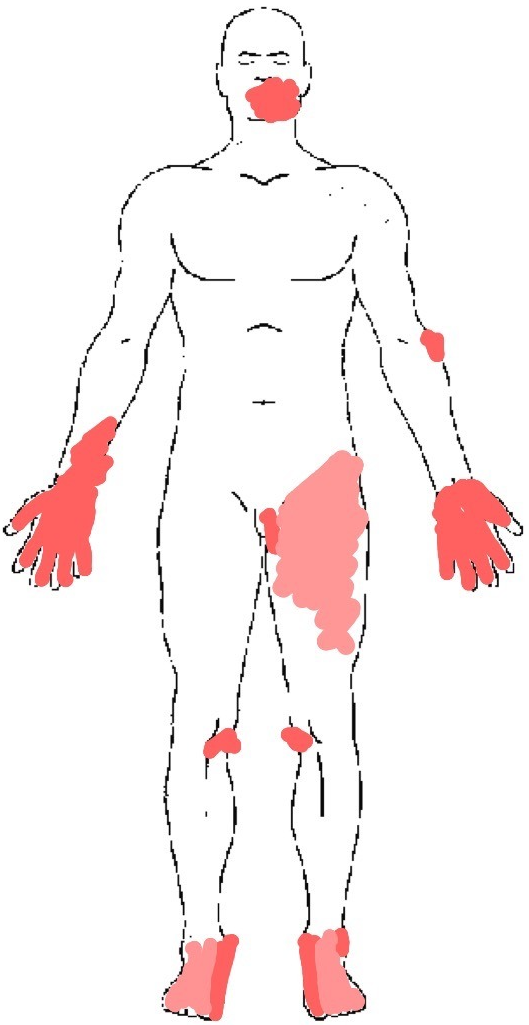}
\includegraphics[height=3cm]{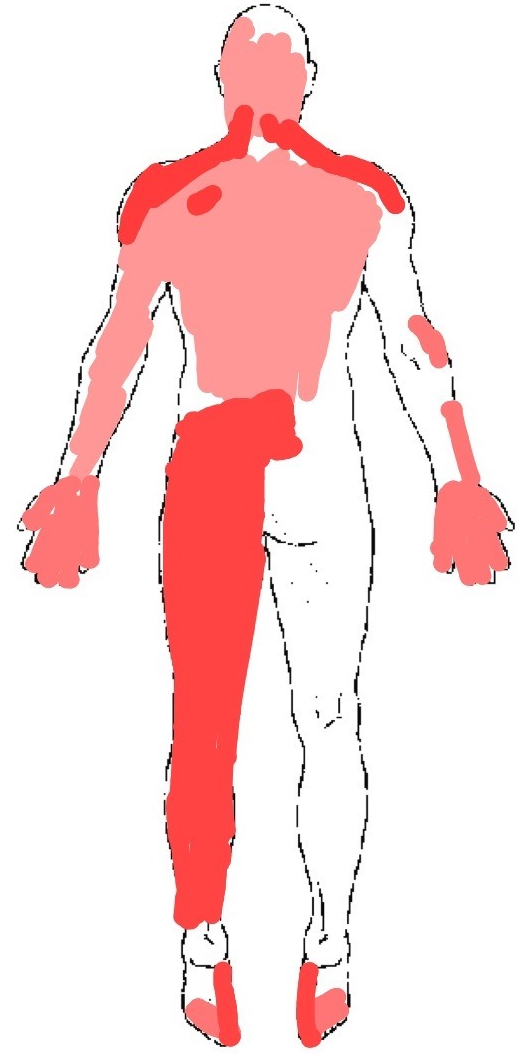}
}}
&\scriptsize \textbf{50 Prd:} 
{ \color{red} Headache;}
{ \color{red}L neck dcf;}
{ \color{red}R neck dcf;}
{\color{blue}Neck dcf;}
{\color{blue}L upper trapezius dcf;}
{\color{blue}R upper trapezius dcf;}
{\color{red}L shoulder dcf;}
{\color{blue}L hand dcf;}
{\color{blue}R hand dcf;}
{\color{red} Interscapular dcf;}
{\color{blue} Lumbago;}
{\color{red} Lateral abdominal dcf;}
{\color{blue}L groin dcf;}
{\color{red}L side thigh dcf;}
{\color{red}R side thigh dcf;}
{\color{blue}L calf dcf;}
{\color{blue}L back thigh dcf;}
{\color{red}L crest of the ilium dcf;}
{\color{red}R crest of the ilium dcf;}
{\color{red}R foot arch dcf;}
{\color{red}L toe joint dcf;}
{\color{red}R  toe joint dcf;}
{\color{red}L medial elbow dcf;}
{\color{blue}L ankle dcf;}
{\color{blue}R ankle dcf;}
{\color{red} L foot arch dcf;}
{\color{red} L PFS;}
{\color{red} L dorsal knee dcf;}
{\color{blue}R medial knee dcf;}\\
&\scriptsize
{\color{blue}L C2 Rdc;
R C2 Rdc;
L C3 Rdc;
R C3 Rdc;
L C4 Rdc;
R C4 Rdc;
R C6 Rdc;
L C6 Rdc;
L C7 Rdc;
R C7 Rdc;
L L5 Rdc; 
R L5 Rdc;
L S1 Rdc;
R S1 Rdc;}
{ \color{red} DLI C2-C3;
DLI C3-C4;
DLI C5-C6;
DLI C6-C7;}
{\color{blue} DLI L4-L5;
DLI L5-S1;}
{ \color{red} OB;}
\\
&\scriptsize \textbf{50 GT:} { \color{red} L back headache; R back headache; }{\color{blue}Neck dcf;} { \color{red} L jaw dcf; } {\color{blue} L upper trapezius dcf; R upper trapezius dcf;} { \color{red} L arm dcf; R arm dcf; L lateral elbow dcf; R lateral elbow dcf; L hand joint dcf; R hand joint dcf; } {\color{blue} L hand dcf; R  hand dcf; } { \color{red} L thumb dcf; R thumb dcf; L finger dcf; R  finger dcf;} {\color{blue}Lumbago;  L groin dcf; L  back thigh dcf;  L calf dcf; L medial knee dcf; L ankle dcf; R ankle dcf; } { \color{red}R  medial knee dcf; R big toe dcf; L  big toe dcf; }\\
&\scriptsize	{\color{blue} L C2 Rdc; R C2 Rdc; L C3 Rdc; R C3 Rdc; L C4 Rdc; R C4 Rdc; L C5 Rdc; R C5 Rdc; L C6 Rdc; R C6 Rdc; L C7 Rdc; R C7 Rdc; } {\color{red}  L L4 Rdc;}  {\color{blue} L L5 Rdc; R L5 Rdc; L S1 Rdc; R S1 Rdc;}	{\color{red} Craniocervical joint injury; } { \color{red}  DLI C4-C5; DLI L3-L4; } {\color{blue} DLI L4-L5; DLI L5-S1; }
\\
\hline
\raisebox{-0.5cm}{\multirow{4}{*}{\includegraphics[height=3cm]{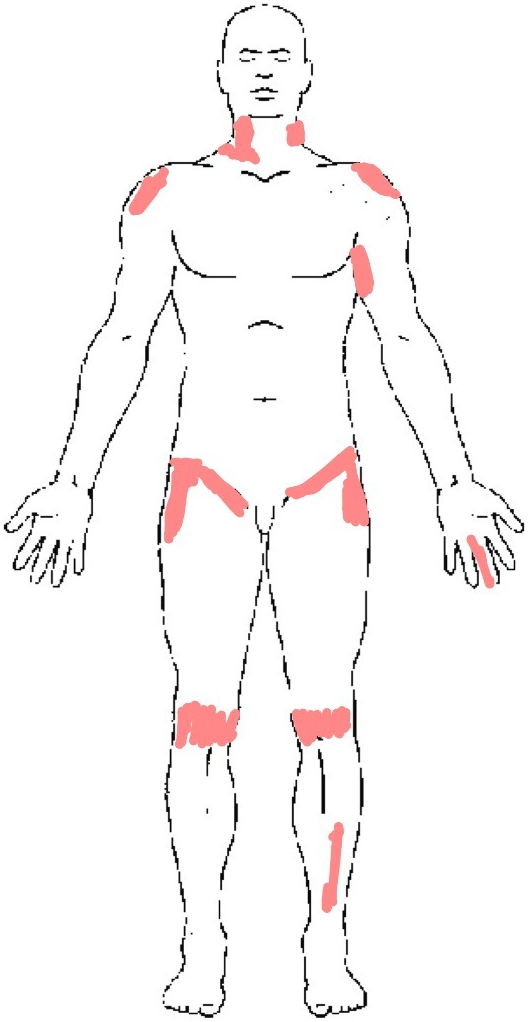}
\includegraphics[height=3cm]{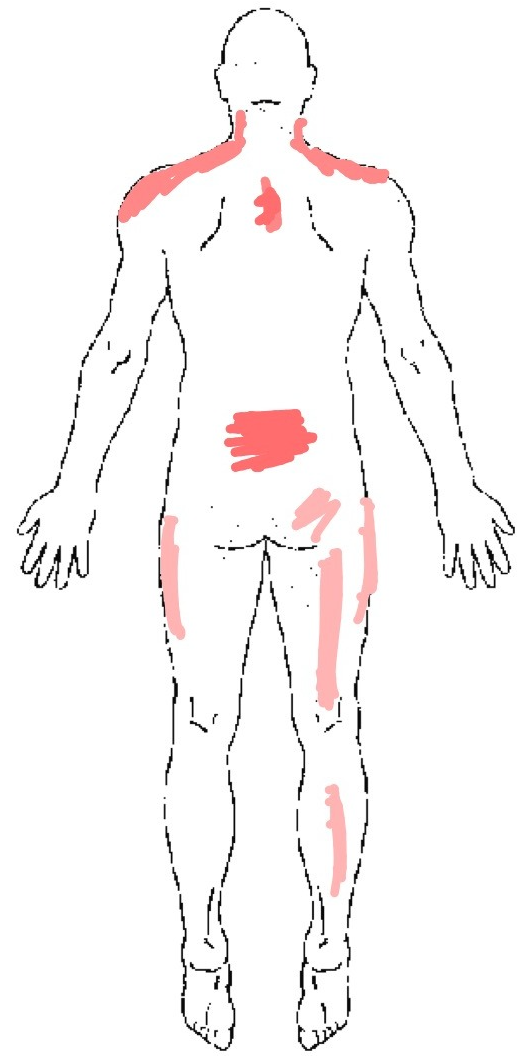}
}}
&\scriptsize \textbf{36 Prd:}
{\color{blue} L neck dcf;} {\color{blue} L shoulder impingement;} {\color{blue} R shoulder impingement;} {\color{blue} L  shoulder dcf;}{\color{blue} R shoulder dcf;} {\color{red} L upper trapezius dcf;} {\color{red} R upper trapezius dcf;} {\color{blue}Lumbago;}{\color{blue}L crest of the ilium dcf;} {\color{blue}R crest of the ilium dcf;} {\color{blue}L adductor tendonitis;} {\color{red}R  back thigh  dcf;} {\color{blue}L PFS;} {\color{blue}R PFS;} {\color{red}R  calf dcf;} {\color{red}L back thigh  dcf;} {\color{red}R anterior knee dcf;}{\color{red}Coccydynia;} {\color{red}L anterior knee dcf;} {\color{red}R medial knee  dcf;}
\\ 
&
\scriptsize
{\color{red}L C4 Rdc;} {\color{red}R C4 Rdc;} {\color{red}L C6 Rdc;} {\color{blue}L C7 Rdc;} {\color{blue}L L5 Rdc;} {\color{blue}R L5 Rdc;} {\color{blue}R S1 Rdc;} {\color{red}L S1 Rdc;} {\color{red}L S2 Rdc;} {\color{red}R S2 Rdc;} {\color{red}DLI C3-C4;} {\color{blue}DLI C5-C6;} {\color{blue}DLI C6-C7;} {\color{blue}DLI L4-L5;}{\color{blue}DLI L5-S1;} {\color{red}DLI S1-S2;}
\\
&\scriptsize \textbf{27 GT:} {\color{blue}L  neck dcf; } {\color{red} R  neck dcf; }{\color{blue}L shoulder impingement; R shoulder impingement; L  shoulder dcf; R  shoulder dcf;; }{\color{red} Interscapular dcf;} {\color{blue}L PFS; R PFS;  Lumbago; L crest of the ilium  dcf; R crest of the ilium  dcf;  L adductor tendonitis; } {\color{red} R adductor tendonitis; R sciatica; L shin discomfort; R side thigh dcf;	}
\\
&\scriptsize {\color{red}  L C5 Rdc; R C5 Rdc; }{\color{blue} L C7 Rdc; L L5 Rdc; R L5 Rdc; R S1 Rdc;	DLI C5-C6; DLI C6-C7;DLI L4-L5; DLI L5-S1; }\\
\hline
\multirow{4}{*}{\includegraphics[height=3cm]{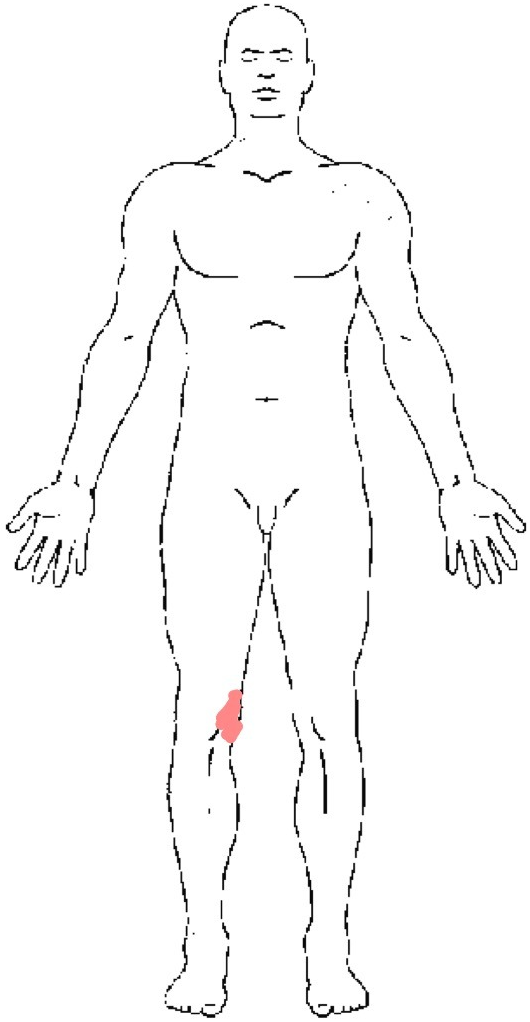}
\includegraphics[height=3cm]{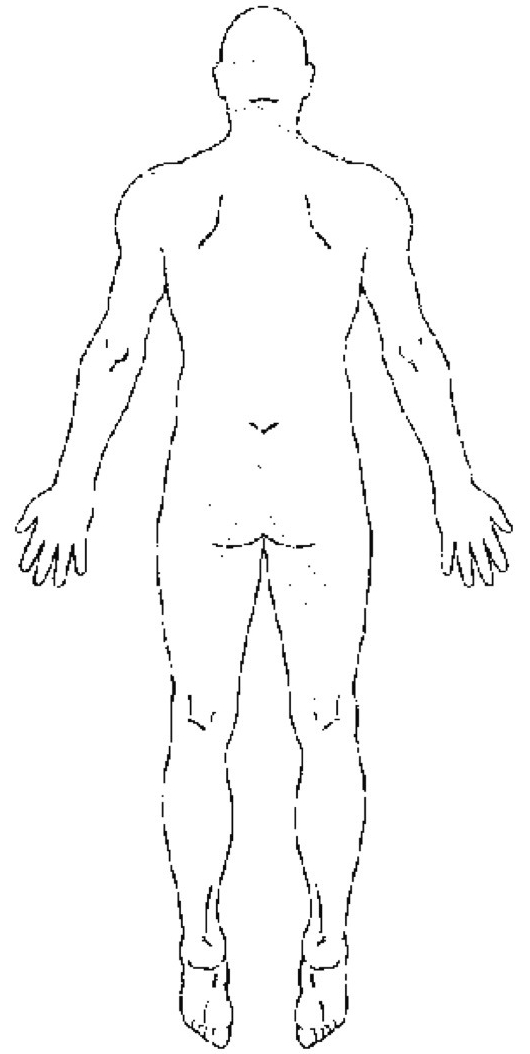}
}
&\scriptsize \textbf{5 Prd:} \color{red} L PFS; R PFS; \\
&\scriptsize  \color{red} R L5 Rdc; DLI L4-L5; L L5 Rdc;\\[5pt]
&\scriptsize \textbf{2 GT:}  \color{red} R Medial knee joint dcf;	\\
&	\scriptsize  \color{red} R Medial meniscus; \\[20pt]
\hline
\end{tabular}
\caption{\footnotesize Prediction examples: The left column shows the input discomfort drawing. The right column shows the predicted diagnostic labels using IBTM after \textbf{Prd:} and the ground truth diagnostic labels given by medical experts after \textbf{GT:}. The number of diagnostic labels is indicated in front of  \textbf{Prd:}  and \textbf{GT:}.Correctly predicted labels are marked in blue, while the wrong ones are marked in red. All labels are given in the order of symptom, pattern and pathophysiology. Rdc stands for Radiculopathy and bcf stands for discomfort in the table.}
\label{tab:PredictionExp}
\vspace{-15pt}
\end{table}
\vspace{-5pt}
Table \ref{tab:PredictionExp} shows typical examples of test results. 
We found that reasonable diagnostic labels can be suggested using IBTM. 
The first  example in Table \ref{tab:PredictionExp} shows a case of high prediction accuracy with a small number of predicted labels. 

The second and third example show typical predictions for which the F-Measure  is in the same level as the mean F-measure. The second example produced a large number of diagnostic labels. 50 Prediction labels are generated and 50 ground truth labels are given. Approximately half of the predictions match the ground truth. However, the mismatched labels are reasonable as well. For example, IBTM predicted Headache, L neck discomfort  R neck discomfort, Neck discomfort; while the ground truth gives L back headache and R back headache and  only a general Neck discomfort. This is caused by different levels of specificity.  These could in fact be considered as correct labels under a more systematic labelling level. The same applies to the prediction of  toe joint discomfort while the ground truth only includes big toe discomfort. In the drawing, the big toe and toe joint region are overlapping.  Additionally, interscapular  discomfort is predicted but not named in the ground truth, although it is marked in the drawing. Hence, when many labels are required, missing labels are easily encountered in both medical expert judgement and machine learning systems. 

The third example shows a case where the number of predictions exceeds significantly  the ground truth labels.  This is due to the difficulty to determine the number of  required predictions, where clustering methods can only aid us but can be inaccurate. Mismatched predictions are the result. The predicted label L upper trapezius discomfort and R upper trapezius discomfort  are apparent in the drawing but not included  in the ground truth. Additionally,  nerves that next to each other are hard to differ due to overlapping symptoms. 

The last example demonstrates highly inaccurate predictions, as the patient has a single local problem. This is a rare case in this dataset. IBTM still suffers from imbalanced data.  However, the predicted label PFS is describing knee problems which seems to be a  reasonable prediction. 

In the end, manually judging the predicted labels, $80\%$ of the labels are in fact reasonable. The measurement in Table \ref{tab:PredictionPrf} is a  rather rough measure without considering more fitting metrics. With a systematic  evaluation standard, for example, considering  the predicted label that has a correspondence on the drawing (upper trapezius discomfort  in the third example) as a correct prediction; and considering the labels within a coherent group as correct predictions (back headache is one type of headache), the F-Measure can be easily recomputed around $70\%$. 
Hence, we believe that with more data and more systematic diagnostic labels, machine learning algorithms can achieve high quality diagnostic predictions. We identify this as an important direction of future work. 

\vspace{-5pt}
\subsection{Diagnostic Interpretation Evaluation} 
In this section, we investigate the structure that the model learned as described in the second part of Section \ref{sec:DPIBTM}. In this evaluation, we train IBTM with all available data.  In the predication phase, we provide one diagnostic label as the label modality and generate a  discomfort drawing. Figure \ref{fig:hist} shows examples of generated drawings with top 10  location words plotted with decreasing intensity for less probable areas. The first row shows examples given a symptom label and the second row shows examples given pattern and pathophysiology label. In the first row, we find that IBTM can generate typical drawings for each symptom, however, it does not always  differ between left and right side correctly. This is caused by the large amount of bilateral symptom labels in the data.  In the second row, Figure \ref{fig:hist} (d) shows a very typical case of left side L4 nerve radiculopathy and (e) and (f) show  typical drawings of DLI L4-L5 and DLI C6-C7. This means that the model is able to learn diagnostic patterns automatically. With a large amount of data, this information can  potentially be  used to help humans to differentiate between different factors in the diagnostic process. 
\vspace{-5pt}
\begin{figure}[h]
\centering
\subfigure[Interscapular discomfort]{
\includegraphics[height=3.5cm]{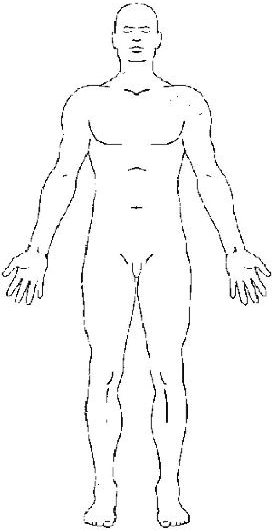}
\includegraphics[height=3.5cm]{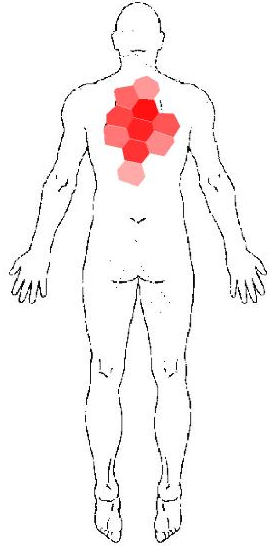}
}
~~~~
\subfigure[Left shin discomfort]{
\includegraphics[height=3.5cm]{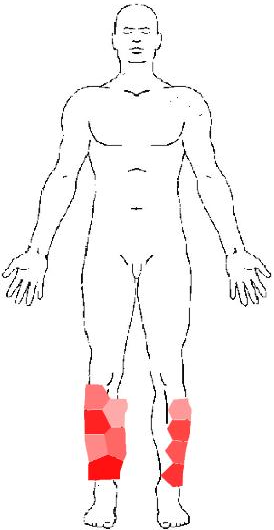}
\includegraphics[height=3.5cm]{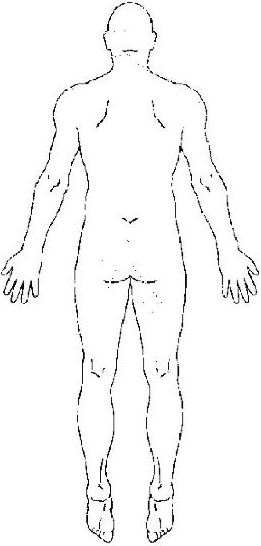}
}
~~~~
\subfigure[Right shin discomfort]{
\includegraphics[height=3.5cm]{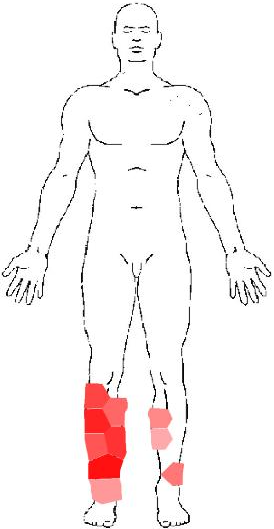}
\includegraphics[height=3.5cm]{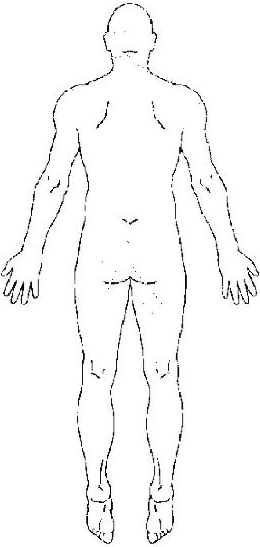}
}

\subfigure[Left L4 Radiculopathy]{
\includegraphics[height=3.5cm]{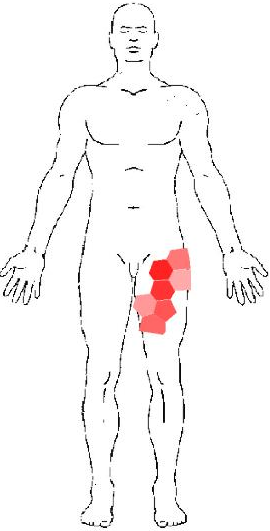}
\includegraphics[height=3.5cm]{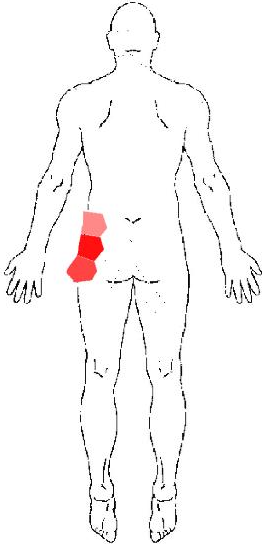}
}
~~~~
\subfigure[DLI L4-L5]{
\includegraphics[height=3.5cm]{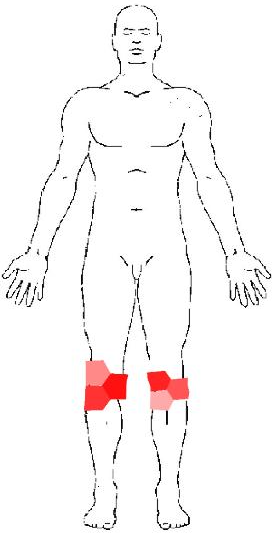}
\includegraphics[height=3.5cm]{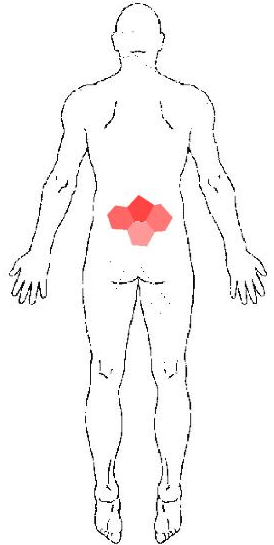}
}
~~~~
\subfigure[DLI C6-C7]{
\includegraphics[height=3.5cm]{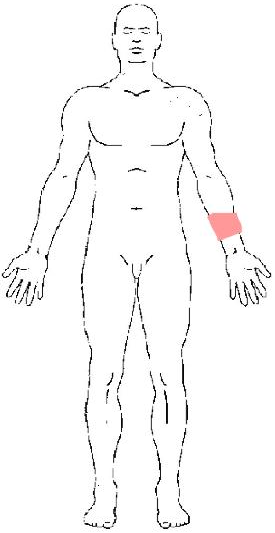}
\includegraphics[height=3.5cm]{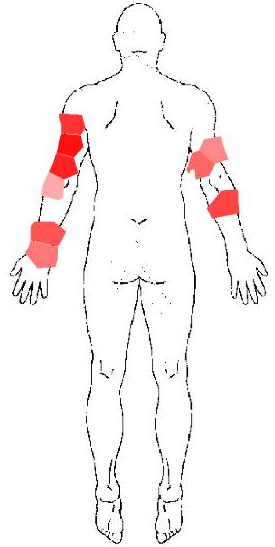}
}
\caption{\footnotesize Generated discomfort drawings given a diagnostic label}
\label{fig:hist}
\end{figure}
\vspace{-5pt}

%% file: discussion.tex
\section{Discussion}
In this paper, we used IBTM for automated assessment of discomfort drawings. A dataset containing real-world discomfort drawings and corresponding diagnostic labels was collected. Reasonable diagnostic predictions were found in the experiments. 
This  preliminary work on this application area shows a promising research direction. We will continue to enlarge and refine the dataset and improve the model. 
At the same time, we will investigate how to present machine learning results to real-life health care personnel. 
We believe that  applying machine learning for diagnostic prediction on  discomfort drawings may have a significant impact on the health care system. It may lead to decision support systems that can help health care personnel to increase effectiveness and precision in diagnosis and treatment of patients.